\documentclass{article}

 \usepackage[final, nonatbib]{neurips_2020}
\usepackage{graphicx,wrapfig}
\usepackage[utf8]{inputenc} 
\usepackage[T1]{fontenc}    
\usepackage{hyperref}       
\usepackage{url}            
\usepackage{booktabs}       
\usepackage{amsfonts}       
\usepackage{nicefrac}       
\usepackage{microtype}      
\usepackage{stackengine}

\newcommand\blfootnote[1]{%
  \begingroup
  \renewcommand\thefootnote{}\footnote{#1}%
  \addtocounter{footnote}{-1}%
  \endgroup
}

\title{Is It a Plausible Colour?\\ UCapsNet for Image Colourisation}

\author{
  Rita Pucci\\
  Department of Computer Science\\
  University of Udine\\
  \texttt{rita.pucci@uniud.it}\\
  \And
    Christian Micheloni\\
  Department of Computer Science\\
  University of Udine\\
  \texttt{Christian.Micheloni@uniud.it} \\
    \And
    Gian Luca Foresti\\
  Department of Computer Science\\
  University of Udine\\
  \texttt{gianluca.foresti@uniud.it} \\
  \And
  Niki Martinel\\
  Department of Computer Science\\
  University of Udine\\
  \texttt{niki.martinel@uniud.it} \\

}

\begin{document}

\maketitle


\begin{abstract}
Human beings can imagine the colours of a grayscale image with no particular effort thanks to their ability of semantic feature extraction. Can an autonomous system achieve that? Can it hallucinate plausible and vibrant colours?
This is the colourisation problem.
Different from existing works relying on convolutional neural network models pre-trained with supervision, we cast such colourisation problem as a self-supervised learning task.
We tackle the problem with the introduction of a novel architecture based on Capsules trained following the adversarial learning paradigm.
Capsule networks are able to extract a semantic representation of the entities in the image but loose details about their spatial information, which is important for colourising a grayscale image.
Thus our UCapsNet structure comes with an encoding phase that extracts entities through capsules and spatial details through convolutional neural networks.
A decoding phase merges the entity features with the spatial features to hallucinate a plausible colour version of the input datum.
Results on the ImageNet benchmark show that our approach is able to generate more vibrant and plausible colours than exiting solutions and achieves superior performance than models pre-trained with supervision.
\end{abstract}

\section{Introduction}
\begin{wrapfigure}{R}{20em}
\centering
\includegraphics[width=\linewidth]{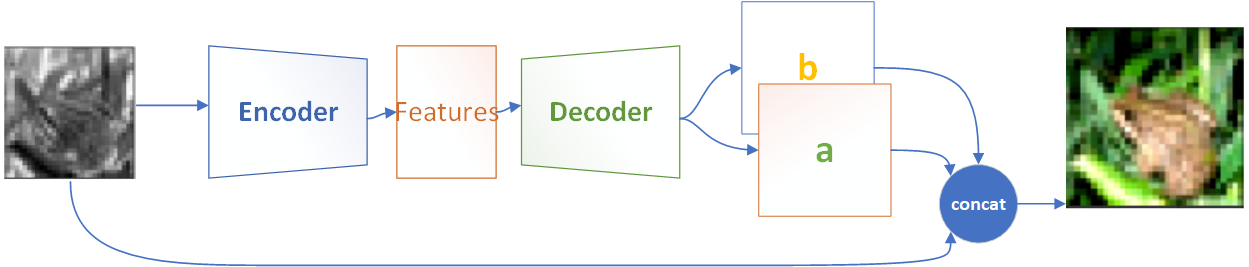}
\caption{Exemplar model structure: the grayscale input image is goes into an encoded that extracts features of the entities present in the image. The encoded representation is then decoded to predict the channels responsible of colourisation.}
\label{fig:baseColourisation}
\end{wrapfigure}

In colourisation, an observer is supposed to know which colours to add to a grayscale image to make it accurate and realistic. Willing to deal with colourisation with an autonomous model, we aim to obtain a realistic/plausible colourisation for a grayscale image by the means of models based on statistical dependencies between the semantic of the objects in the image and its grayscale texture. Existing works mainly rely on Convolutional Neural Networks (CNNs) to obtain de-saturated image colours~\cite{cheng2015deep,charpiat2008automatic,deshpande2015learning}, with a few exceptions~\cite{larsson2016learning,iizuka2016let,zhang2016colorful, isola2017image} achieving plausible colourisation for a limited set of samples through auxiliary losses.  We believe that such unsatisfactory results are due to the limited abilities of CNNs in understanding the semantic information of an entity in relation to its parts \cite{sabour2017dynamic}. Indeed, it is a matter of fact that colours are closely related to the semantics of a scene. This motivates us to introduce an approach that grasps semantic information and generates a plausible colourisation that will potentially fool an observer. We cast the colourisation problem as a self-supervised learning task (Fig.\ref{fig:baseColourisation}) under the adversarial learning framework. Specifically, we introduce the \textbf{UCapsNet} architecture that leverages the convolutional operators to get spatial features and combines these with capsules~\cite{he2016deep,sabour2017dynamic} for the semantic features extraction process. A capsule consists of a group of neurons that collaborate to generate anactivity vector. The activity vector represents the probability of an entity existence in the image (length of the vector) and its instantiated parameters/features (direction of the vector). We entangle a strong interaction between the spatial features and the entities information through an encoder-decoder solution that exploits skip connections. Such a colour generator shows a spatial and semantic comprehension of the entities present in the images and a consistent ability of colourisation that is optimised through the adversarial min-max game with a discriminator. Results on ImageNet show that UCapsNet outperforms existing solutions (some of which exploits pre-trained models).

\section{Related Work}
Colourisation algorithms propose to extract data useful to map the grayscale channel onto the coloured ones.
The literature can be clustered into non-parametric and parametric methods.
The former exploits image analogies based on a multiscale autoregression between two images, where colours are transferred onto the input image taking into consideration analogous regions of the other images~\cite{hertzmann2001image,welsh2002transferring,gupta2012image,liu2008intrinsic,zhang2017real}. The latter extracts knowledge from training datasets, it can be later applied for colourisation, in this paper our approach embrace the same idea. In~\cite{deshpande2015learning}, authors apply the LEARCH framework to balance pixelwise accuracy and spatial error and scene labelling to pick the appropriate scene specific regressor. In~\cite{larsson2016learning}, a fully connected pre-trained VGG16 model incorporates semantic features and a colour histogram framework to predict colourisation. Taking up the idea of collaboration between pre-trained layers,~\cite{NTIRE} introduces a model of Dense Blocks structured following the UNet~\cite{ronneberger2015u} architecture. With the idea of considerign semantic and spatial information, in~\cite{iizuka2016let}, four component collaborate to extract classification information along side chrominance features by the means of CNNs layers that merge local patch information with the global one obtained from the entire image. In~\cite{zhang2016colorful},  a model based on CNNs layers is trained to learn a quantised representation of colours per each pixel. In ~\cite{isola2017image}, a Generative Adversarial Model (GAN) is trained to obtained a complete autonomous colourisation system. In~\cite{ozbulak2019image}, a pre-trained VGG19~\cite{simonyan2014very} model is added to the CapsNet~\cite{sabour2017dynamic} architecture and fine-tuned for colourisation. Differently from all such methods, UCapsNet (i) explores the collaboration between CNNs and capsules layers to merge spatial and semantic features (ii) with the aim of generating a plausible colorization (iii) by means of a self-supervised learning solution that does not hinge on supervised pre-training. 

\begin{wrapfigure}{L}{23em}
    \centering
    \includegraphics[scale = 0.4]{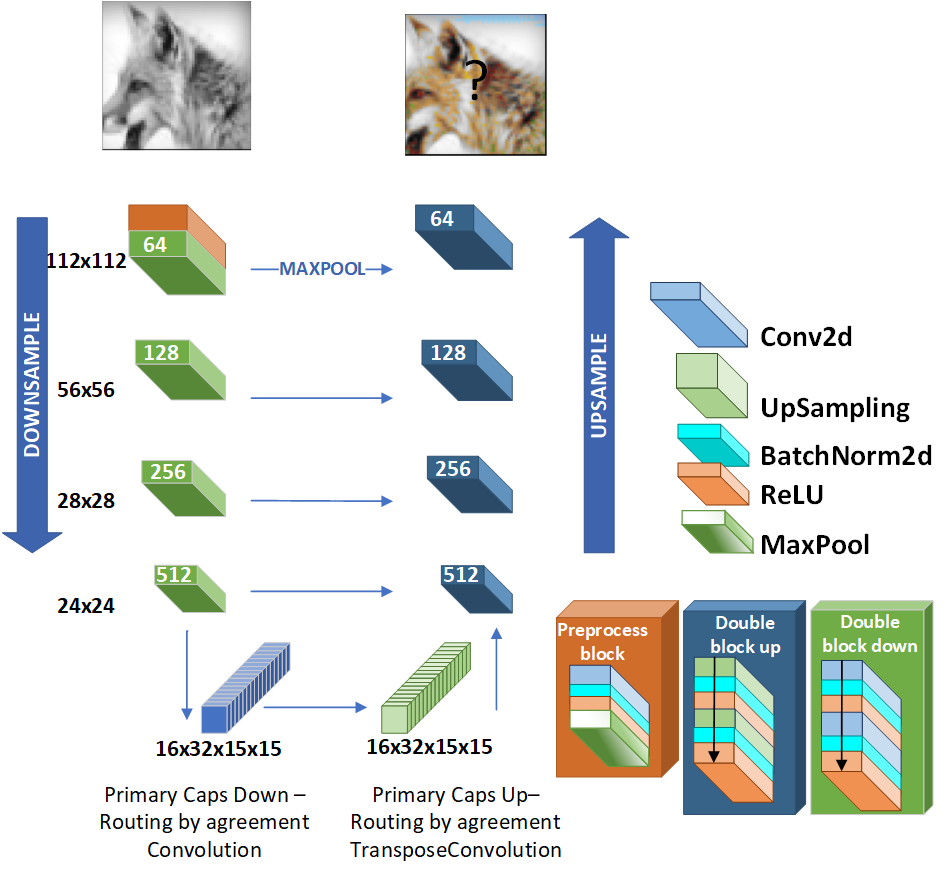}
    \caption{UCapsNet architecture. Numbers within each box denote the number of feature maps, while values on the sides indicate the corresponding spatial dimensions.
    Best viewed in colours.}
    \label{fig:architecture}
\end{wrapfigure}

\section{Approach}
\textbf{Image preprocessing:} we followed the previous literature~\cite{zhang2016colorful,iizuka2016let,isola2017image,larsson2016learning,ozbulak2019image} and considered images in the CIELab colourspace.
This colourspace considers three channels: \textit{L} (luminance), \textit{a} and \textit{b} (chrominance).
It is a perceptually linear colourspace as it establishes a mapping between the colours in the Euclidean space and the colours as perceived by humans (i.e., unlike the RGB colourspace, CIELab is designed to approximate human vision).
\\
\textbf{Colour quantization:} we take the idea of colours quantization from \cite{zhang2016colorful}.
The \textit{ab} channels 2D-space is quantized into $Q=313$ values in gamut using a grid size of 10.
The aim is to learn a mapping $G : \mathbf{I} \mapsto \mathbf{Z}$ with $\mathbf{I}$ being the grayscale input image and $\textbf{Z} \in [0, 1]^{H \times W \times Q}$ corresponding to the per-pixel probability distribution over the $Q$ colours.
\\
\textbf{Overall Architecture:}
The proposed network architecture is shown in Fig.~\ref{fig:architecture}.
The main computational blocks are the \textit{double block down} (green box) and the \textit{double block up} (blue box).
We followed a "U"-shaped architecture presenting a downsample (encoding) phase that applies the \textit{double block down} to reduce the spatial definition of the input by extracting multiple features at different levels.
After each block, the output is sent to the next block and to the corresponding \textit{double block up} in the deconding phase (i.e., through a skip-connection).
With this we aim to extract and maintain the spatial information of the images which is crucial for colourisation.
\\
\textbf{Details\footnote{\url{https://github.com/Riretta/Colourisation_w_Capsules}}:} The \textit{downsample phase} starts with a preprocess block (orange box in Fig.~\ref{fig:architecture}) that consists of a $2D$ convolution having a $7\times 7$ kernel and stride of $2\times 2$ with
Batch Normalization (BN) and Rectified Linear Unit (ReLU) activation function.
A max pooling layer is used in the skip-connection to ensure that the output spatial dimension is compliant with the upsampling correspondent layer.
All the subsequent operators are \textit{double block down} layers.
Each consists of two $2D$ convolutional layers with $3\times 3$ kernels and $1\times1$ strides.
BN and ReLU layers follow each convolutional one.
Each \textit{double block down} doubles the number of input features maps while halving their spatial resolution.
At the end of downsampling, we obtain $512$ features maps of size $24\times 24$.
These are fed to the capsules (Primary Caps layers) to extract information about the entities present in the input image.
The 16 blue capsules in \textit{Primary Caps Down} consist of convolutional layers applied to extract features used during the Routing by agreement, as described in \cite{sabour2017dynamic}, to extract high level entities information. The entities information extracted are fed to the \textit{Primary Caps Up} that performs an UpSampling operation representing the beginning of the \textit{upsample phase}.
These \textit{Primary Caps up} consists of 16 capsules. The output of the\textit{Primary caps up} is the input of the first \textit{double block up}. 
This block receives the entities features from the Primary Caps layers and the spatial features from the skip-connected \textit{double block down}.
The upsampling phase proceeds with three double blocks up that reconstruct the predicted quantization of colours of the input image.
Each \textit{double block up} consists of two UpSampling layers activated with BN and ReLU layers. 
Given a $224\times 224$ input, the network outputs a matrix $\hat{\mathbf{Z}} \in \mathbb{R}^{56\times 56\times 313}$.
\\
\textbf{Colourisation:} starting from $\hat{\mathbf{Z}}$, an RGB image is reconstructed by first applying an inverse mapping from the $Q=313$ values in gamut to the $ab$ coordinates, then by the concatenation of the results with the input grayscale datum.

\textbf{Quantization loss:} to enforce generation of vibrant colours, we adopted the multinomial cross entropy loss proposed in~\cite{zhang2016colorful}:
\begin{equation}
\label{Lcl}
\mathcal{L}_{cl}(\hat{\textbf{Z}}, \textbf{Z}) = - \sum_{q}\textbf{Z}_{h,w,q} log(\hat{Z}_{h,w,q})  
\end{equation}
that compares the predicted $\hat{\textbf{Z}}$ quantization against the ground truth $\textbf{Z}$.
\\
\textbf{Training procedure:} to train our UCapsNet model, we considered a self-supervised learning procedure where, given an image, the input to the model is the corresponding $L$ channel and the ground-truth are the remaining $ab$ ones. We followed a Generative Adversarial Network (GAN)~\cite{goodfellow2014generative,goodfellow2016nips} where the Markovian discriminator (PatchGAN) \cite{isola2017image} has been considered as the discriminator $D(\cdot)$ and and our proposed UCapsNet defined the generator $G(\cdot)$. The optimisation problem considers a combination of $\mathcal{L}_{GAN} (G, D)=E_{y}[log D(y)]+ E_{x,z}[log(1-D(G(x, z))]$ , the traditional loss $\mathcal{L}_{L1}(G) = E_{x,y,z}[\|y-G(x, z)\|_1]$, and the $\mathcal{L}_{cl}$ defined in equation (\ref{Lcl}):
\begin{equation}
    G^* = \arg\stackunder{min}{G}\stackunder{max}{D}\mathcal {L}_{GAN}(G, D) +\mathcal{L}_{L1}(G)+\mathcal{L}_{cl}
\end{equation}

\section{Experimental Results}
\textbf{Dataset:} to validate the performance of our approach we have considered the ImageNet dataset~\cite{ILSVRC15}.
The training set has been used for model fitting.
For a fair comparison with other methods, from the ImageNet evaluation set, we considered the same 1000 samples adopted in~\cite{vitoria2020chromagan} for colourisation.
\\
\textbf{Optimisation:} the Adam optimiser~\cite{kingma2014adam} with a learning rate of $2\times10^{-5}$ has been used to minimise the objective losses for both the generator and the discriminator.
We trained our model for 100 epochs with a batch size of 32 using the PyTorch framework and an NVIDIA Titan Xp.
\\
\textbf{Model variants:} we present results obtained with UCapsNet trained with and without the GAN framework (in the latter case, only the quantization loss is considered). We also report on the performance obtained by learning to directly predict the $ab$ channels (UCapsNet AB(GAN)), thus predicting $\hat{\textbf{Z}_{ab}}\in\mathcal{R}^{56\times 56\times 2}$. In such a case the model is optimised with the mean squared error loss in place of $\mathcal{L}_{cl}$.
\begin{wrapfigure}{R}{25em}
    \centering
    \includegraphics[width=\linewidth]{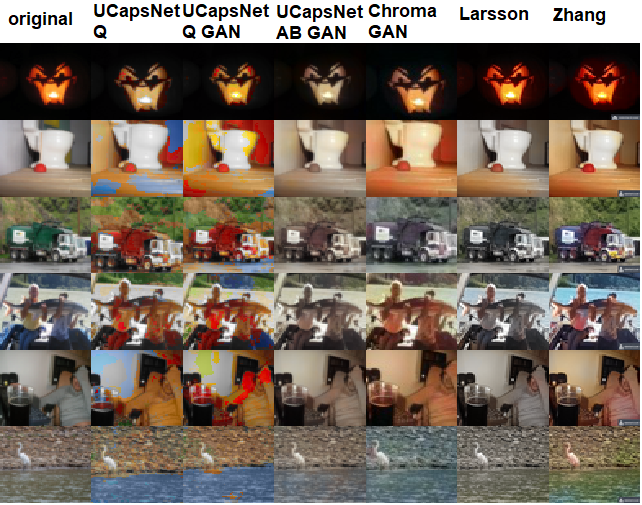}
    \caption{Qualitative colourisation ImageNet performance. Comparison between the results  obtained with different UCapsNet variants and the existing solutions.}
    \label{fig:results}
\end{wrapfigure}

\textbf{Performance:}
To visually assess the performance of our approach we computed the results in Fig.~\ref{fig:results}. These show that UCapsNet has promising colourisation capabilities. Comparable colourisation performance with more complex structures are obtained even if UCapsNet we do not apply mechanisms to deepen by the means of pre-trained model (e.g.,~\cite{vitoria2020chromagan}). Results obtained by Larsson~\cite{larsson2016learning} and by UCapsNet AB(GAN) tend to output muted colours that tend to be more "brownish" than the vibrant colours obtained by UCapsNet Q(GAN) and Zhang~\cite{zhang2016colorful}. UCapsNet is also able to reconstruct plausible information by adding vivid colours that better match the entity in the image. In fact, the second, third, and sixth rows demonstrate that the colours generated by our solution are more plausible than the ones predicted by other methods. The inconsistent splotches obtained by UCapsNet Q are generally well addressed by means of the adversarial approach (second and fifth rows). Also, colour boundaries are not perfectly detected by UCapsNet Q and Larsson~\cite{larsson2016learning}. To have an overall analysis of our performance, we followed a common approach~\cite{larsson2016learning,vitoria2020chromagan} and computed the peak signal-to-noise ratio (PSNR) of the predicted \textit{ab} images with respect to the ground truth and compared to those obtained for other fully automatic methods.
The results shown in Tab.~\ref{PSNR} demonstrate that our model has the best overall performance.
This results indicate a better colourisation performance throughout the test dataset.
\begin{wraptable}{r}{5.0cm}
  \caption{PSNR (dB) results on 1000 ImageNet validation samples.}
  \label{PSNR}
  \begin{tabular}{lcc}
    \toprule
    \cmidrule(r){1-2}
    Model   & PSNR (dB)\\
    \midrule
    Isola et al~\cite{isola2017image}& 21.57\\
    Larsson et al~\cite{larsson2016learning} & 24.93\\
    Zhang et al~\cite{zhang2016colorful} & 22.04\\
    Vitoria et al~\cite{vitoria2020chromagan} & 25.57\\
      \midrule
    Ours (Q)  &           28.43\\
    Ours (Q GAN)&       28.42 \\
    Ours (AB GAN)&        28.57\\
    \bottomrule
  \end{tabular}
\end{wraptable}
 \section{Conclusion}
In this paper, the UCapsNet architecture based on Capsule Network (CapsNet) is designed for the image colourisation problem under a self-supervised learning setup.
The proposed method is a fully automatic, end-to-end, deep model that explores the collaboration between spatial features extracted by convolutional layers and entity features extracted by Capsule layers. This model exploits the generative adversarial network framework to learn a plausible colourisation model. Experiments with the ImageNet dataset show that UCapsNet has superior performance than exiting works considering pre-trained models. Future works will focus on designing a deeper UCapsNet architecture and on investigating the application of our colourisation approach as a pretext task.
\textbf{Acknowledgement} we thank Harry S. Rugg for proofreading.

\bibliographystyle{abbrv}
  \bibliography{egbib}
\blfootnote{NeurIPS 2020 Workshop: Self-Supervised Learning - Theory and Practice}
\end{document}